\documentclass[letterpaper, 10 pt, conference]{ieeeconf}  

\IEEEoverridecommandlockouts                              

\usepackage{graphicx} 
\usepackage{tabularx}
\usepackage{stfloats}
\usepackage{url}
\usepackage{cite}
\usepackage{algorithm}
\usepackage{algorithmic}
\usepackage{amsfonts}
\usepackage{amsmath}
\usepackage{array,booktabs,arydshln}
\usepackage{bbm}
\usepackage[table]{xcolor}  
\usepackage{colortbl}  
\usepackage{subfig}
\usepackage{booktabs}
\usepackage{float}
\definecolor{lightblue}{HTML}{D2E4FC}
\newcommand{\best}[1]{\cellcolor{lightblue}\textbf{#1}}
\newcommand{\negval}[1]{\textcolor{red}{#1}}  

\usepackage{flushend}

\definecolor{green}{RGB}{11,155,13}

\title{\LARGE \bf
GACL: Grounded Adaptive Curriculum Learning with \\Active Task and Performance Monitoring }

\author{Linji Wang$^{1}$, Zifan Xu$^{2}$, Peter Stone$^{2, 3}$, and Xuesu Xiao$^{1}$ 
\thanks{
        $^{1}$Department of Computer Science, George Mason University {\tt\small \{lwang44, xiao\}@gmu.edu}
        $^{2}$Department of Computer Science, The University of Texas at Austin {\tt\small zfxu@utexas.edu, pstone@cs.utexas.edu}
        $^{3}$Sony AI
        }%
}

\begin{document}

\maketitle
\thispagestyle{empty}
\pagestyle{empty}

\begin{abstract}

Curriculum learning has emerged as a promising approach for training complex robotics tasks, yet current applications predominantly rely on manually designed curricula, which demand significant engineering effort and can suffer from subjective and suboptimal human design choices. While automated curriculum learning has shown success in simple domains like grid worlds and games where task distributions can be easily specified, robotics tasks present unique challenges: they require handling complex task spaces while maintaining relevance to target domain distributions that are only partially known through limited samples. 
To this end, we propose Grounded Adaptive Curriculum Learning (GACL\footnote{\url{https://github.com/linjiw/GACL}}), a framework specifically designed for robotics curriculum learning with three key innovations: (1) a task representation that consistently handles complex robot task design, (2) an active performance tracking mechanism that allows adaptive curriculum generation appropriate for the robot's current capabilities, and (3) a grounding approach that maintains target domain relevance through alternating sampling between reference and synthetic tasks. 
We validate GACL on wheeled navigation in constrained environments and quadruped locomotion in challenging 3D confined spaces, achieving 6.8\% and 6.1\% higher success rates, respectively, than state-of-the-art methods in each domain.

\end{abstract}

\section{Introduction}

\label{sec:intro}
Curriculum learning has shown promises in training robots for complex tasks such as navigating through highly constrained environments or maintaining quadruped locomotion across challenging terrain~\cite{xiao2024autonomous, atanassov2024curriculum}. However, current applications of curriculum learning in robotics face a fundamental challenge: they predominantly rely on manually designed curricula, which demand significant engineering effort and can suffer from subjective, suboptimal design choices. For example, in quadruped locomotion tasks~\cite{atanassov2024curriculum}, roboticists must carefully design progressive stages from basic jumping skills to complex obstacle traversal and manually define success metrics and progression conditions at each stage.

While automated curriculum learning (ACL) can reduce manual design effort, current approaches have primarily focused on simple domains like grid worlds and games where tasks can be easily specified~\cite{portelas2020automatic}. Recent advances like PAIRED~\cite{dennis2020emergent} and CLUTR~\cite{azad2023clutr} have improved upon basic ACL by introducing teacher agents for task generation. However, these methods face fundamental limitations when applied to complex robotics problems: their stateless teachers cannot leverage performance history, they lack domain grounding mechanisms, and their simplified task representations cannot handle complex robotics applications.

\begin{figure}[t]
    \centering
    \includegraphics[width=\columnwidth]{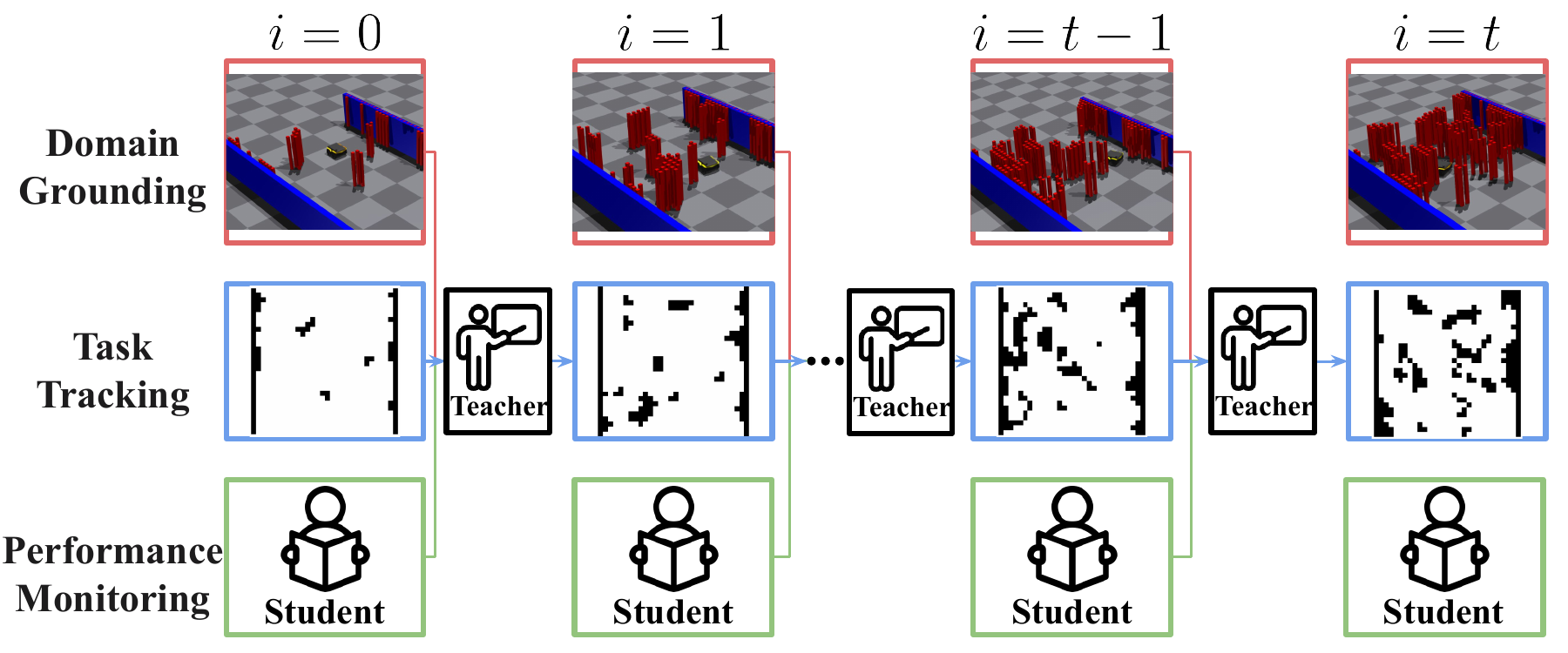}
    \caption{
    GACL creates an adaptive sequence of robot tasks with progressively increasing difficulty through target domain grounding, task tracking, and performance monitoring.  
    }
    \label{fig:overview}
    \vspace{-19pt}
\end{figure}
More fundamentally, robotics tasks present a unique challenge overlooked by existing ACL approaches: open-ended curricula often move towards tasks that are not relevant to target deployment domains. Every robotics problem has an final deployment domain (like household environments for home robots) whose complete distribution is unknown and can only be partially observed through limited samples. While generative models can approximate these distributions, robotic curriculum learning requires generating tasks that not only match the target distribution but also adapt to the robot's current learning capabilities. This creates a critical balance: the curriculum must generate diverse training scenarios that challenge the robot during learning while ensuring these scenarios remain relevant to the target domain where the robot will eventually operate.

To address these challenges, we propose Grounded Adaptive Curriculum Learning (GACL), a framework specifically designed for robotics curriculum learning.  This paper introduces three key innovations: (1) a task representation that consistently handles complex robot task design, (2) an active performance tracking mechanism that allows adaptive curriculum generation appropriate for the robot's current capabilities, and (3) a grounding approach that maintains target domain relevance through alternating sampling between reference and synthetic tasks (Fig.~\ref{fig:overview}).

We validate GACL on two challenging robotics domains: wheeled robot navigation in highly constrained environments~\cite{xu2023benchmarking} and quadruped locomotion in  confined 3D spaces~\cite{xu2024dexterous}. GACL demonstrates consistent improvements in both domains, achieving 6.8\% and 6.1\% higher success rates, respectively, compared to state-of-the-art curriculum learning methods.

\section{Related Work}
\label{sec::related}

Curriculum learning, originally inspired by the way humans acquire skills, has been explored in robotics to structure learning from increasingly complex tasks \cite{bengio2009curriculum}. Early work often relied on carefully hand-crafted curricula, where experts define stage-by-stage task progression, such as denser obstacle distributions~\cite{perille2020benchmarking}, increased elevation changes~\cite{xu2024reinforcement}, and more complex locomotion strategies~\cite{atanassov2024curriculum}. While these manual curricula can improve learning, they require significant domain expertise, often introducing subjective biases that can lead to suboptimal training efficiency and final performance~\cite{narvekar2020curriculum}. To address these limitations, researchers have turned to ACL to reduce human effort and automatically adapt training to improve efficiency and performance.


ACL seeks to offload the burden of manually designing a sequence of tasks to a teacher agent that observes the student agent’s learning progress and adaptively proposes new tasks~\cite{portelas2020automatic}. A primary goal in ACL is to challenge the student agent appropriately—neither too easy nor too hard—thereby improving sample efficiency and robustness. However, most ACL work were developed in relatively simple domains, such as grid worlds and basic physics simulations~\cite{campbell2023automatic}, where tasks can be represented in low-dimensional spaces.

Recent studies on Unsupervised Environment Design (UED) highlight more advanced teacher-student frameworks. One prominent example is PAIRED \cite{dennis2020emergent}, which uses a regret-based objective to push the student agent toward increasingly challenging environments. CLUTR \cite{azad2023clutr} builds upon PAIRED by integrating a learned latent representation (using variational autoencoders \cite{kingma2013auto}) for task generation, offering more efficient task sampling. Despite these advances, existing UED methods still exhibit notable limitations when applied to complex robotics tasks: (1) The task space remains relatively simple in CLUTR, which focuses on low-dimensional representations. In contrast, real-world robotics tasks often require more complex inputs—such as multi-channel images for navigation maps or multi-layer 3D terrain for legged locomotion—that cannot be easily captured by low-dimensional encoding; and (2) Most teacher implementations, including CLUTR’s, operate in a stateless, multi-armed bandit fashion. This stateless approach restricts the teacher’s capacity to model the nuanced relation between student performance and task complexity, which is particularly problematic for robotics scenarios involving high-dimensional sensor data and intricate physical dynamics.

One of the most critical yet underexplored challenges in applying ACL to robotics is ensuring that the automatically generated tasks remain relevant to real-world deployment scenarios. While open-ended task generation can foster broader policy generalization, unconstrained approaches risk producing training scenarios that deviate significantly from the environments in which the robot will operate. This mismatch can lead to wasted training effort on unrealistic tasks and failed task execution when facing real-world settings.
Therefore, grounding the curriculum to the robot's target deployment domain with existing real-world experiences is therefore crucial. Yet, most existing ACL methods, including PAIRED and CLUTR, do not balance the need for realistic task distributions and pushing the agent's capabilities. Specifically, while CLUTR uses VAE-based representations, it lacks the stateful tracking and domain grounding essential for robotics applications.

In summary, existing curriculum learning solutions face three central challenges in robotics: (1) scaling automated curricula to high-dimensional, complex tasks; (2) monitoring student performance to adapt tasks efficiently; and (3) preserving domain relevance while still pushing the agent's capabilities. 
In the next section, we introduce GACL, a framework designed explicitly to overcome these hurdles by combining richer task representations, active performance monitoring, and a grounding mechanism that ensures each generated task reflects real-world deployment conditions.

\section{Approach}
\label{sec:approach}

\begin{figure}[t]
    \centering
    \includegraphics[width=\columnwidth]{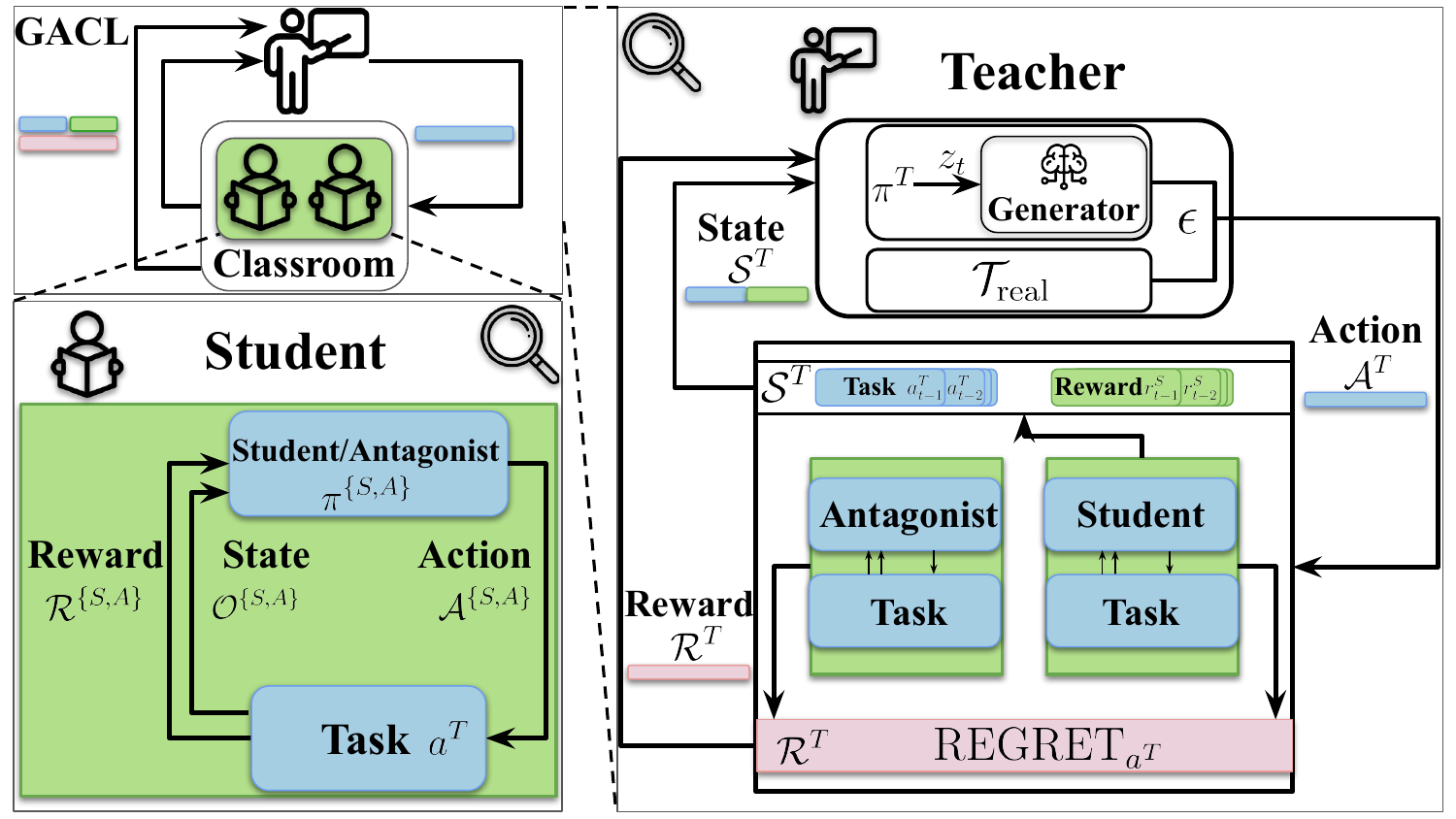}
    \caption{Overview of the Dual-Agent GACL Framework (top left): student POMDP (bottom left) and teacher MDP (right).}
    \label{fig:gcl_approach}
    \vspace{-15pt}
\end{figure}

We propose GACL, a dual-agent framework designed for adaptive curriculum learning in complex robotic tasks while maintaining target domain relevance. As depicted in Fig.~\ref{fig:gcl_approach}, GACL involves two interacting components:

\begin{itemize}
    \item  A Partially Observable Markov Decision Process (POMDP) for the student agent ($\cdot^S$) learning the task.
    \item A fully informed Markov Decision Process (MDP) for the teacher agent ($\cdot^T$) generating a curriculum of tasks.
\end{itemize}

\subsection{Dual-Agent (PO)MDP}
\label{subsec:problem_formulation}

\subsubsection{Student Agent POMDP}
The student agent operates in a POMDP defined as a tuple $\mathcal{M}^S = \langle \mathcal{S}^S, \mathcal{A}^S, \mathcal{O}^S, \mathcal{T}^S, \Omega^S, \mathcal{R}^S, \gamma^S \rangle$, where:
\begin{itemize}
    \item $\mathcal{S}^S$ is the state space of the robotic task,
    \item $\mathcal{A}^S$ is the robot action space,
    \item $\mathcal{O}^S$ is the robot observation space,
    \item $\mathcal{T}^S: \mathcal{S}^S \times \mathcal{A}^S \rightarrow \mathcal{S}^S$ is the POMDP transition function,
    \item $\Omega^S: \mathcal{S}^S \times \mathcal{A}^S \rightarrow \mathcal{O}^S$ is the robot observation function,
    \item $\mathcal{R}^S: \mathcal{S}^S \times \mathcal{A}^S \times \mathcal{S}^S \rightarrow \mathbb{R}$ is the robot reward function based on task execution performance, and
    \item $\gamma^S \in [0, 1]$ is the student POMDP's discount factor.
\end{itemize}
The Student agent's goal is to learn a policy $\pi^S: \mathcal{O} \rightarrow \mathcal{A}$ that maximizes the expected cumulative reward in the partially observable task environment generated by the teacher.
\subsubsection{Teacher Agent MDP}
In contrast to the student's POMDP, the teacher agent operates in an MDP defined as $\mathcal{M}^T = \langle \mathcal{S}^T, \mathcal{A}^T, \mathcal{T}^T, \mathcal{R}^T, \gamma^T \rangle$, where:
\begin{itemize}
    \item $\mathcal{S}^T$ is the teacher state space, consisting of the comprehensive history of tasks and student performances,
    \item $\mathcal{A}^T$ is the teacher action space representing all possible tasks that can be assigned to the student,
    \item $\mathcal{T}^T: \mathcal{S}^T \times \mathcal{A}^T \rightarrow \mathcal{S}^T$ is the MDP transition function,
    \item $\mathcal{R}^T: \mathcal{S}^T \times \mathcal{A}^T \times \mathcal{S}^T \rightarrow \mathbb{R}$ is the teacher reward function based on student performance, and
    \item $\gamma^T \in [0, 1]$ is the discount factor for the teacher's MDP.
\end{itemize}
$s_t^T \in \mathcal{S}^T$ at time $t$ is defined as $s_t^T = \{(a_i^T, r_i^S)\}_{i=0}^{t-1}$, where $a_i^T \in \mathcal{A}^T$ is the $i$ th task assigned by the teacher and $r_i^S$ is the student's performance (reward) for task $i$.

\subsection{Grounded Adaptive Curriculum Learning (GACL)}
\label{subsec:gcl}
GACL employs a hierarchical structure where a fully informed teacher agent guides the learning of a student agent, resembling a classroom setting (Fig.~\ref{fig:gcl_approach}, top left). In this metaphor, the teacher (right side of the figure) not only determines the curriculum (i.e., the tasks) but also monitors the student’s progress and compares it against an antagonist agent (bottom left). This design mirrors real-world educational scenarios, where teachers have comprehensive knowledge of both course material and student performance. Leveraging this vantage point, GACL introduces three key innovations for complex robotics tasks: 
\emph{a latent generative model} that consistently encodes and reconstructs high-dimensional task environments, 
\emph{active performance tracking} to adaptively challenge the student at its evolving skill level, 
and \emph{domain relevance maintenance} via alternating between synthetic tasks and limited real-world references. Together, these elements ensure that GACL provides diverse yet realistic training scenarios, prevents the learning process from drifting into irrelevant tasks, and aligns the curriculum with the student’s incremental improvements.

\subsubsection{Task Representation via Latent Generative Model}
\label{subsubsec:latent_model}
We employ a Variational Autoencoder (VAE) to learn a continuous latent space $\mathcal{Z}$ for high-dimensional robotic tasks, trained on a partially known real-world task set $\mathcal{T}_{\mathrm{real}}$. The encoder-decoder architecture compresses and reconstructs complex environments (e.g., 2D navigation maps or 3D terrain heightmaps). During curriculum learning, the teacher agent controls task generation by sampling latent vectors $\mathcal{Z}$, which the VAE decoder maps to concrete tasks. This VAE is pre-trained offline on $\mathcal{T}_{\mathrm{real}}$ prior to the main GACL training loop, providing a stable latent representation throughout the curriculum learning process.

\subsubsection{Student and Antagonist Agents}
\label{subsubsec:student_agent}
The student agent learns to perform a task using a reinforcement learning algorithm (e.g., PPO \cite{schulman2017proximal}) in the partially observable environment generated by the teacher. Its objective is to maximize the expected cumulative reward:
\begin{equation}
J^S(\pi^S_{\theta^S}) = \mathbb{E}_{\tau^S \sim \pi^S_{\theta^S}}\left[\sum_{t=0}^{T} {(\gamma^S)}^t r^S_t\right],
\label{eqn::J^S}
\end{equation}
where $\tau^S$ is a trajectory sampled from the student policy $\pi^S_{\theta^S}$, parameterized by $\theta^S$.
To guide curriculum generation and evaluate the student's progress, we introduce an antagonist agent, following the flexible regret setting from PAIRED~\cite{dennis2020emergent}. The antagonist is trained with the same observability and hyperparameters as the student, sharing the same objective function (Eqn.~\eqref{eqn::J^S}), with $J^A$ and $\pi^A_{\theta_A}$ as the antagonist's objective and policy respectively.

\subsubsection{Teacher Agent}
\label{subsubsec:teacher_agent}
The teacher agent in GACL monitors two key elements to design adaptive curricula: student performance and task history. 
The teacher maintains a history of student rewards, $\{r_i^S\}_{i=0}^{t-1}$, stored in its state $s_t^T$ to monitor student performance. By incorporating these past performance metrics, the teacher can dynamically adjust the curriculum to match the student’s evolving skill level.
The teacher also retains a history of previously assigned tasks $\{a_i^T\}_{i=0}^{t-1}$ in $s_t^T$ and exploits the latent space $\mathcal{Z}$ learned by the VAE trained on partially known real-world data to produce teacher action based on the latent embedding $z_t$. Based on both student performance and task history, the teacher generates new tasks using $z_t$ for the student by sampling from this latent space using the VAE decoder $G: \mathcal{Z} \rightarrow \mathcal{A}^T$, which maps latent vectors to concrete tasks.
The teacher then aims to maximize the expected cumulative regret:

\begin{equation}
    J^T(\pi^T_{\theta^T}) = \mathbb{E}_{\tau^T \sim \pi^T_{\theta^T}}\left[\sum_{t=0}^{T} {(\gamma^T)}^t \textsc{Regret}_{a^T_t}(\pi^S_{\theta^S}, \pi^A_{\theta_A})\right], \nonumber
\end{equation}
where:
\begin{itemize}
    \item $\tau^T = (s_0^T, a_0^T, s_1^T, a_1^T, ..., s^T_t)$ is a trajectory in the teacher's MDP, 
    \item $\pi^T_{\theta^T}$ is the teacher policy, parameterized by $\theta^T$, 
    \item $\textsc{Regret}_{a^T_t}(\pi^S_{\theta^S}, \pi^A_{\theta_A}) = V_{a^T_t}(\pi^A_{\theta_A}) - V_{a^T_t}(\pi^S_{\theta^S})$ is the regret for task $a^T_t$, generated by the teacher at $t$, and
    \item $V_{a^T_t}(\cdot)$ is the value function (expected discounted return) of a policy when executing task $a^T_t$.
\end{itemize}
\begin{algorithm}[tbp]
\small
\caption{Grounded Adaptive Curriculum Learning}
\label{alg:gacl}
\begin{algorithmic}[1]
\STATE \textbf{Input:} VAE decoder $G$, initial parameters $\theta^S$, $\theta^A$, and $\theta^T$, learning rates $\eta^S$, $\eta^A$, and $\eta^T$, real-world task set $\mathcal{T}_{\text{real}}$, and grounding probability $\epsilon$
\STATE \textbf{Output:} Trained policies $\pi^S_{\theta^S}$, $\pi^A_{\theta^A}$, and $\pi^T_{\theta^T}$
\STATE Pretrain $G$ with available real-world tasks $\mathcal{T}_{\text{real}}$
\STATE Initialize $\pi^S_{\theta^S}$, $\pi^A_{\theta^A}$, $\pi^T_{\theta^T}$, and $s_0^T = \{\}$
\STATE $t \gets 0$
\WHILE{not converged}

    \STATE ${a_t^T} \gets \begin{cases}
        \text{sample from } \mathcal{T}_{\text{real}}, & \text{with probability } \epsilon, \\
        \pi^T_{\theta^T}(s_t^T), & \text{with probability } 1-\epsilon,
    \end{cases}$
    \STATE Collect student trajectory in task ${a_t^T}$ and compute cumulative reward $r^S_{{a_t^T}}$ 
    \STATE Collect antagonist trajectory in task ${a_t^T}$

    \STATE Compute regret $\textsc{Regret}_{{a_t^T}} \gets V_{{a_t^T}}(\pi^A_{\theta^A}) - V_{{a_t^T}}(\pi^S_{\theta^S})$
    \STATE Update $s_{t+1}^T \gets s_t^T \cup \{({a_t^T}, r^S_{{a_t^T}})\}$
    \STATE $\pi^S \gets \pi^S + \alpha^S \nabla_{\pi^S} J^S(\pi^S)$
    \STATE $\pi^A \gets \pi^A + \alpha^A \nabla_{\pi^A} J^A(\pi^A)$
    \STATE $\pi^T \gets \pi^T + \alpha^T \nabla_{\pi^T} J^T(\pi^T)$
 
    \STATE $t \gets t + 1$
\ENDWHILE
\RETURN $\pi^S_{\theta^S}$, $\pi^A_{\theta^A}$, and $\pi^T_{\theta^T}$
\end{algorithmic}
\end{algorithm}
\subsubsection{Maintaining Domain Relevance via Alternating Sampling}
\label{subsec:balancing_tasks}
Ensuring that learned policies remain aligned with the target deployment domain is crucial for real-world applicability. To this end, GACL interleaves \emph{reference tasks} from $\mathcal{T}_{\mathrm{real}}$ with \emph{synthetic tasks} sampled by the teacher from its latent space by augmenting teacher action $a_t^T$:

\begin{equation}
    {a_t^T} = \begin{cases}
        \text{sample from } \mathcal{T}_{\text{real}}, & \text{with probability } \epsilon, \\
        \pi^T_{\theta^T}(s_t^T), & \text{with probability } 1-\epsilon, \nonumber
    \end{cases}
\end{equation}
where $\epsilon \in [0,1]$ controls the probability of selecting a real-world reference task at each step. 

Algorithm~\ref{alg:gacl} summarizes the GACL training loop. 
We first pretrain the VAE on the available real-world tasks $\mathcal{T}_{\mathrm{real}}$ (line~3). The teacher agent then alternates between sampling reference tasks (\(\epsilon\) probability) and generating new tasks via $\pi^T_{\theta^T}$ (line~7). Both student and antagonist collect trajectories in the selected task (lines~8--9), after which the teacher computes regret and updates its state (lines~10--11). Finally, all three policies (student, antagonist, and teacher) update their parameters according to their respective objectives (lines~12--14). This process continues until convergence.

\section{Experiments}
\label{sec:experiments}
In this section, we evaluate GACL against three baselines: Base RL (no curriculum), Manual (expert) Curriculum Learning (Manual CL), and CLUTR \cite{azad2023clutr}. We test on two challenging domains: BARN Navigation~\cite{perille2020benchmarking, xu2023benchmarking} and Quadruped Locomotion~\cite{xu2024dexterous}. We also conduct ablation studies by removing each of our three main components to analyze their contributions. Here, we describe our experimental setup, evaluation metrics, results, and curriculum visualizations.

\subsection{Experimental Setup}
\label{subsec:exp_setup}

We conduct experiments in two challenging robotics domains: (1) BARN Navigation in highly constrained environments, and (2) Quadruped Locomotion in confined 3D spaces. Both tasks are implemented in simulation, using partial domain knowledge to ground curriculum generation. 
Specifically, we use procedurally generated environments to represent the final target deployment domain, but the generation procedure is unknown to the teacher agent, which needs to ground its curriculum through only partial knowledge.  
Below, we detail each domain’s setup, instantiate the student and teacher agent, and summarize the key hyperparameters.

\subsubsection{BARN Navigation}

\paragraph{Environment}
We adopt the BARN map generator to produce highly constrained 2D navigation maps, featuring narrow corridors and cluttered layouts~\cite{perille2020benchmarking}. For this experiment, we simulate 128 parallel environments in IsaacGym \cite{makoviychuk2021isaac} to accelerate data collection and RL training. 
\paragraph{Student Agent}
The student POMDP's observation space $\mathcal{O}^S$ includes the robot's position and orientation, 270° field-of-view, 720-dimensional LiDAR scans, and the relative goal orientation; action space $\mathcal{A}^S$ comprises continuous linear and angular velocities; and reward function $\mathcal{R}^S$ encourages progress towards the goal while penalizing collisions and excessive time.
\paragraph{Teacher Agent}
The teacher MDP's action space $\mathcal{A}^T$ is the task latent space, with $s_t^T = \{(a_{t-1}^T, r_{t-1}^S)\}$ tracking the most recent task and performance.

\subsubsection{Quadruped Locomotion}

\paragraph{Environment}
We simulate a quadruped robot moving in confined 3D environments with rugged terrain and overhanging obstacles and maintain 128 parallel instances.

\paragraph{Student Agent}
The student POMDP's observation space $\mathcal{O}^S$ encompasses the robot's base pose, joint states (angles and velocities), partial terrain, and ceiling heightmaps; action space $\mathcal{A}^S$ consists of continuous torque commands for each leg; and reward function $\mathcal{R}^S$ incentivizes stable forward motion and penalizes falls, collisions, and undue time expenditure.

\paragraph{Teacher Agent}
The teacher MDP's action space $\mathcal{A}^T$ encodes both terrain and ceiling configurations in the VAE's latent space, with state representation $s_t^T = \{(a_{t-1}^T, r_{t-1}^S)\}$ following the same structure as navigation.

\subsubsection{Hyperparameters}
Table \ref{tab:hyperparameters} summarizes the key hyperparameters used in our experiments.

\begin{table}[tbp]
\centering
\caption{Hyperparameters for GACL.}
\label{tab:hyperparameters}
\begin{tabular}{lll}
\specialrule{1.2pt}{0pt}{0pt}
\textbf{GACL Parameter} & \textbf{Value} & \\
\midrule
Parallel Environments  & 128 & \\
Latent Task Dimension  & 32  & \\
Training Epochs        & 5000 & \\
\midrule
\textbf{RL Parameter}  & \textbf{Teacher} & \textbf{Student}\\
\midrule
Learning Rate          & 1e-4  & 3e-4 \\
PPO Epoch              & 10    & 5 \\
Discount Factor        & 0.99  & 0.99 \\
\specialrule{1.2pt}{0pt}{0pt}
\end{tabular}
\vspace{-15pt}
\end{table}

\subsection{Evaluation Metrics}
\label{subsec:eval_metrics}

For evaluation, we employ a comprehensive set of metrics to assess both student performance and curriculum difficulty.

\paragraph{Student Performance.}

In the BARN Navigation domain, we measure (i)~\emph{Success Rate} (the percentage of trials that reach the goal without collisions), (ii)~\emph{Navigation Progress} (the proportion of the path traversed before failure), (iii)~\emph{Average Steps} per successful episode, (iv)~\emph{Average Reward} (summing forward progress and collision penalties), and (v)~\emph{Average Speed} (m/s).
For Quadruped Locomotion, we track (i)~\emph{Success Rate} (no falls before reaching the endpoint), (ii)~\emph{Distance Traveled} (m), (iii)~\emph{Footstep Efficiency} (ratio of time steps with at least three legs in ground contact), and (iv)~\emph{Average Reward} reflecting gait stability and forward progress.

\paragraph{Curriculum Difficulty.}
To quantify task complexity, we define a domain-specific difficulty score. 
For BARN Navigation, 
\(
D_{\mathrm{nav}} = \alpha \,(\mathrm{path\ length}) \;-\; \beta\,(\mathrm{clearance}),
\)
where $\mathrm{path\ length}$ denotes the shortest collision-free path in the map and $\mathrm{clearance}$ is the minimum distance to obstacles along that path.
For Quadruped Locomotion, 
\(
D_{\mathrm{loc}} = \gamma \,(\mathrm{terrain\ slope}) \;+\; \delta\,(\mathrm{obstacle\ density}),
\)
where $\mathrm{terrain\ slope}$ measures elevation changes and $\mathrm{obstacle\ density}$ measures how rough and bumpy the terrain is, including the number of uneven features.

\section{Results and Discussion}
\label{sec:results}
\subsection{Main Results}
\label{subsec:main_results}
\begin{table*}[tb]
\centering
\caption{Performance Comparison on BARN Navigation and Quadruped Locomotion (mean $\pm$ std).
$\uparrow$ indicates higher is better; $\downarrow$ indicates lower is better.}
\label{tab:main_results}
\resizebox{\textwidth}{!}{%
\begin{tabular}{lcccccccc}
\specialrule{1.2pt}{0pt}{0pt}
& 
\multicolumn{4}{c}{\textbf{BARN Navigation}} & 
\multicolumn{4}{c}{\textbf{Quadruped Locomotion}} \\
\cmidrule(lr){2-5} \cmidrule(lr){6-9}
\textbf{Method}
& \textbf{Success(\%)}$~\uparrow$ 
& \textbf{Prog(\%)}$~\uparrow$ 
& \textbf{Steps}$~\downarrow$
& \textbf{Reward}~$\uparrow$ 
& \textbf{Success(\%)}~$\uparrow$
& \textbf{Dist.(m)}~$\uparrow$ 
& \textbf{Footstep Eff.}~$\uparrow$ 
& \textbf{Reward}~$\uparrow$ \\
\midrule
Base RL        
& 76.16 $\pm$ 4.47 
& 64.06 $\pm$ 2.38 
& \best{36.41} $\pm$ 0.15 
& 18.36 $\pm$ 0.84 
& 72.34 $\pm$ 3.26
& 4.41  $\pm$ 1.57 
& 78.6  $\pm$ 2.5
& 16.74 $\pm$ 0.82
\\
Manual CL      
& 76.83 $\pm$ 5.02 
& 66.17 $\pm$ 2.47 
& 36.65 $\pm$ 0.22 
& 19.19 $\pm$ 1.17 
& 74.12 $\pm$ 4.89
& 4.67  $\pm$ 1.94 
& 79.3  $\pm$ 2.7
& 17.02 $\pm$ 1.13
\\
CLUTR          
& 76.67 $\pm$ 2.74 
& 66.52 $\pm$ 4.23 
& 36.70 $\pm$ 0.28 
& 18.39 $\pm$ 0.65 
& 74.65 $\pm$ 2.31
& 4.76  $\pm$ 1.66
& 80.2  $\pm$ 3.1
& 17.28 $\pm$ 0.95
\\
\textbf{GACL (Ours)}
& \best{81.85} $\pm$ 2.51
& \best{68.89} $\pm$ 2.77
& 36.99 $\pm$ 0.57
& \best{19.45} $\pm$ 0.77
& \best{79.21} $\pm$ 2.54
& \best{5.12}  $\pm$ 1.80
& \best{82.9}  $\pm$ 2.6
& \best{18.41} $\pm$ 0.77
\\
\specialrule{1.2pt}{0pt}{0pt}
\end{tabular}}
\vspace{-10pt}
\end{table*}
Table~\ref{tab:main_results} presents the performance of four methods---Base RL, Manual CL, CLUTR, and GACL---on both the BARN Navigation and Quadruped Locomotion tasks. 

\paragraph{BARN Navigation}
GACL outperforms all baselines in \emph{Success Rate} (81.85\%), \emph{Progress} (68.89\%), and \emph{Average Reward} (19.45), demonstrating superior navigation in complex environments. While Base RL achieves the lowest \emph{Average Steps}, it suffers more collisions due to prioritizing speed over safety. Manual CL shows competitive performance but is limited by its fixed progression schedule, and CLUTR's stateless teacher fails to adapt effectively to the student's developing capabilities.

\paragraph{Quadruped Locomotion}
Similarly, GACL leads in all metrics for the locomotion task: \emph{Success Rate} (79.21\%), \emph{Distance Traveled} (5.12\,m), \emph{Footstep Efficiency} (82.9\%), and \emph{Reward} (18.41). The combination of domain-grounded task tracking and active performance monitoring enables GACL to generate appropriately challenging terrain variations. CLUTR moderately improves over Base RL but is hampered by its lack of domain grounding, while Manual CL provides minor improvements but cannot adapt to the full spectrum of terrain difficulties.

Overall, these results demonstrate GACL's effectiveness in both domains through its adaptive curriculum scheduling based on domain grounding, task tracking, and performance monitoring.

\subsection{Ablation Studies}
\label{subsec:ablation_studies}

We conduct ablation studies to evaluate the contribution of each key component in GACL: (i)~domain grounding, (ii)~task tracking, and (iii)~performance monitoring. Each variant removes one of these components from the full GACL framework:

\begin{itemize}
    \item GACL: The complete GACL framework.
    
    \item GACL \texttt{w/o grounding}: Removes domain grounding. We discard reference task sampling, relying solely on the teacher agent’s synthetic tasks:
    \(
    a_t^T = \pi^T_{\theta^T}(s_t^T).
    \)

    \item GACL \texttt{w/o task}: Removes task tracking. We replace the latent task representation \(a_i^T\) with a random vector \(\xi_i\) in the teacher’s state:
    \(
    s_t^T = \{(\xi_i,\, r_i^S)\}_{i=0}^{t-1},
    \)
    where 
    \(
    \xi_i \sim \mathcal{N}(0,\, I).
    \)

    \item GACL \texttt{w/o performance}: Removes performance monitoring. We replace the student’s reward \(r_i^S\) with a random scalar \(\eta_i\) in the teacher’s state:
    \(
    s_t^T = \{(a_i^T,\, \eta_i)\}_{i=0}^{t-1},
    \)
    where 
    \(
    \eta_i \sim \mathcal{U}(0,\, 1).
    \)
\end{itemize}

Table~\ref{tab:gacl_ablation} shows the \emph{Success Rate} (\%) for each ablation variant in both BARN Navigation and Quadruped Locomotion, along with the difference (Diff.) from the full GACL. The full framework consistently outperforms its ablated variants, confirming that each component---domain grounding (\texttt{w/o grounding}), task tracking (\texttt{w/o task}), and performance monitoring (\texttt{w/o performance})---is essential. Removing domain grounding yields the largest performance drop (\(-5.49\%\) in Navigation and \(-5.11\%\) in Locomotion), showing that partial real-world references are key for maintaining task relevance. Omitting performance monitoring or task tracking likewise impairs curriculum adaptation, highlighting the importance of aligning generated tasks with the robot’s evolving capabilities.

\begin{table}[tbp]
\centering
\caption{Ablation Study: Success Rate (\%) in Navigation and Locomotion.
All differences (Diff.) are relative to the full GACL. Negative values are shown in red.}
\label{tab:gacl_ablation}
\resizebox{\columnwidth}{!}{%
\begin{tabular}{lcccc}
\specialrule{1.2pt}{0pt}{0pt}
& \multicolumn{2}{c}{\textbf{Navigation}} & \multicolumn{2}{c}{\textbf{Locomotion}} \\
\cmidrule(lr){2-3}\cmidrule(lr){4-5}
\textbf{Variant} 
& \textbf{Success(\%)} & \textbf{Diff.} 
& \textbf{Success(\%)} & \textbf{Diff.} \\
\midrule
GACL
  & \best{81.85} & --  
  & \best{79.21} & -- \\
GACL \texttt{w/o grounding}        
  & 76.36 & \negval{-5.49} 
  & 74.10 & \negval{-5.11} \\
GACL \texttt{w/o task}        
  & 79.86 & \negval{-1.99} 
  & 77.31 & \negval{-1.90} \\
GACL \texttt{w/o performance} 
  & 77.69 & \negval{-4.16} 
  & 75.94 & \negval{-3.27} \\
\specialrule{1.2pt}{0pt}{0pt}
\end{tabular}}
\vspace{-10pt}
\end{table}

\subsection{Curriculum Progression}
\label{subsec:curriculum_progression}
\begin{figure*}[tbp]
\centering
\includegraphics[width=\textwidth]{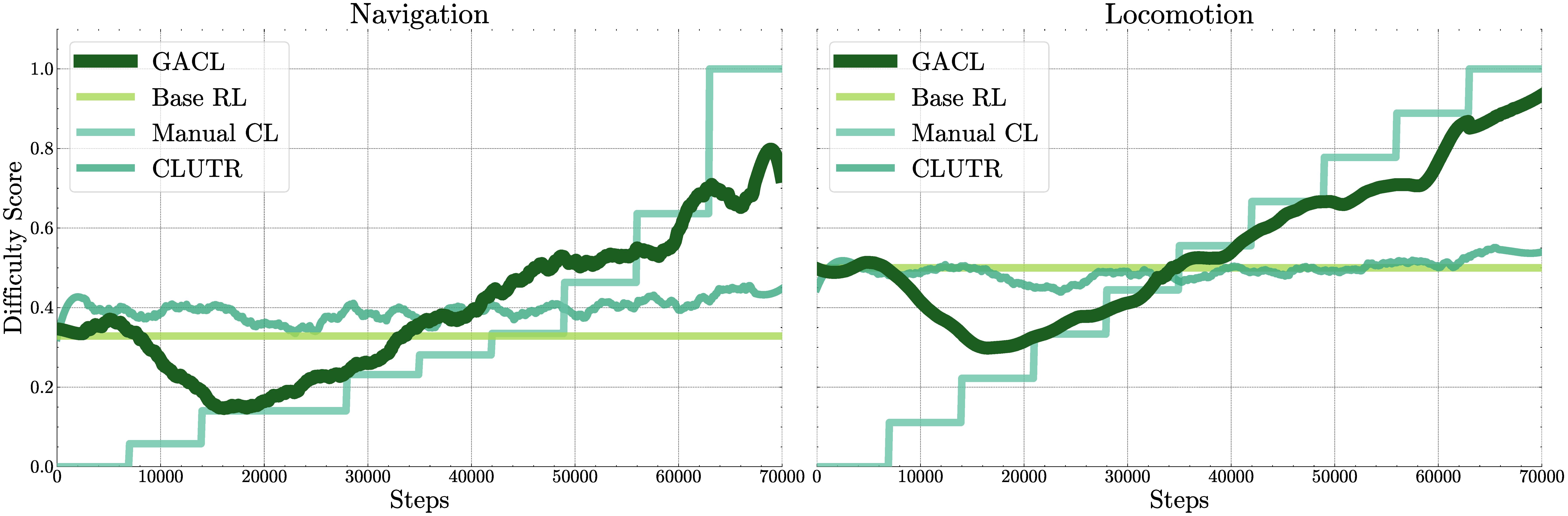}
\caption{Curriculum difficulty trends for BARN Navigation (left) and Quadruped Locomotion (right). 
GACL initially lowers difficulty from a moderate level as it detects the student’s struggles, 
then ramps up complexity once the student becomes more proficient.}
\vspace{-15pt}
\label{fig:difficulty_progression}
\end{figure*}

We visualize how each method evolves task difficulty during training using the heuristic measures described in Section~\ref{subsec:eval_metrics}. Fig.~\ref{fig:difficulty_progression} compares Base RL, Manual CL, CLUTR, and GACL in both the BARN Navigation domain (left) and the Quadruped Locomotion domain (right). The \emph{y}-axis represents the difficulty score (higher is more difficult), and the \emph{x}-axis denotes training steps.

For both tasks, Base RL never adjusts task complexity and thus remains effectively uniform throughout training. Manual CL follows a fixed schedule, steadily increasing difficulty from simpler tasks to harder ones. CLUTR hovers near a middle range of task complexity, showing limited capacity to adapt the curriculum to the student’s evolving skill. In contrast, GACL dynamically adjusts task difficulty based on student learning stages. Early in training, GACL’s difficulty curve starts near a moderate level—reflecting the VAE’s initialization around a mean task distribution—then decreases as the teacher learns to manipulate the latent space to produce simpler tasks for the struggling student. Once the student becomes more proficient, the teacher progressively increases task difficulty, prompting further skill development. This adaptive scheduling showcases the synergy among GACL’s task-awareness, performance tracking, and partial domain grounding, ultimately yielding more effective curricula than competing methods.

\begin{figure}[tb]
    \centering
    \includegraphics[width=0.95\columnwidth]{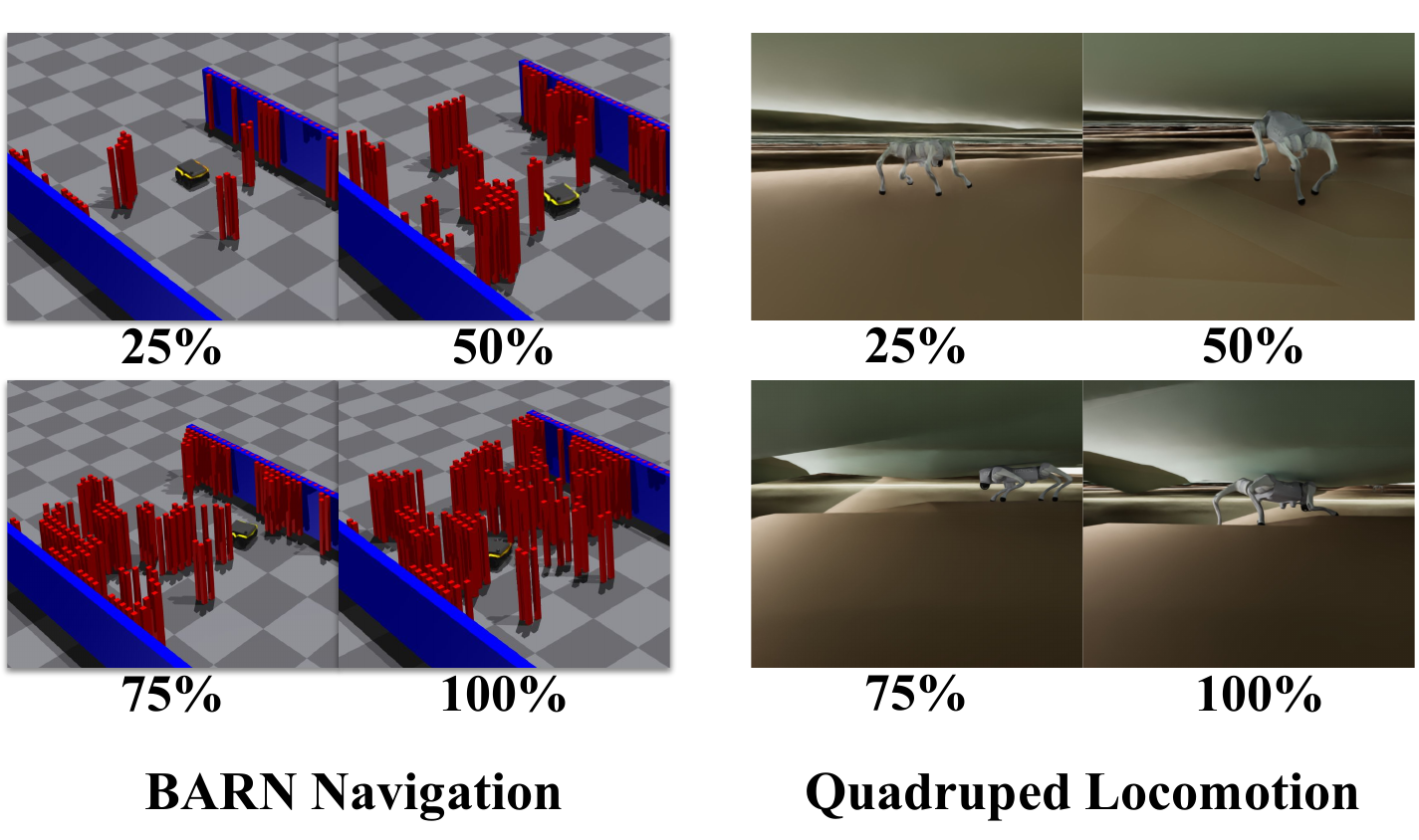}
    \caption{Progression of tasks generated by GACL over training. 
    Left: BARN Navigation maps sparse to complex. 
    Right: Confined 3D terrains for Quadruped Locomotion, increasing in slope irregularities and obstacles.}
    \label{fig:env_vis}
    \vspace{-10pt}
\end{figure}

\label{subsec:qualitative_vis}
Fig.~\ref{fig:env_vis} displays snapshots of the curriculum generated by GACL at four training milestones (25\%, 50\%, 75\%, and 100\% of total training). On the left, BARN Navigation tasks evolve from open, sparsely cluttered layouts to narrower corridors populated with numerous obstacles, mirroring the student’s increasing competence. On the right, Quadruped Locomotion terrain and ceiling shift from gently sloped surfaces to progressively more uneven and intricate terrain and ceiling obstacles, challenging the robot’s growing capabilities. This visual progression illustrates GACL’s adaptive approach to curriculum design, where task difficulty increments as the agent’s proficiency advances.

\subsection{Discussion}
\label{subsec:discussion}

Our experiments demonstrate GACL's effectiveness across two distinct robotics domains with different sensing and control requirements. The ablation studies confirm that all three components---domain grounding, task tracking, and performance monitoring---are essential to GACL's success, as removing any one significantly degrades performance. These results highlight the potential of our unified framework to automatically generate adaptive curricula for complex robot tasks while maintaining focus on relevant target-domain challenges.

\section{Conclusions and Future Work}
\label{sec::conclusions}

This paper introduces GACL, a framework specifically tailored to complex robotics domains by actively maintaining domain relevance, task awareness, and performance tracking. Our experiments on two tasks---wheeled robot navigation in the BARN challenge and quadruped locomotion in confined 3D spaces---demonstrate that GACL consistently outperforms both state-of-the-art automated curriculum methods and carefully designed expert curricula. Notably, GACL achieves 6.8\% and 6.1\% higher success rates in the navigation and locomotion tasks, respectively, compared to SOTA approaches, illustrating its ability to balance structured curricula with flexible adaptation to the robot’s evolving capabilities.
Through our ablation studies, we observed that each GACL component is critical: \textit{(i)} domain grounding prevents the curriculum from drifting toward unrealistic scenarios, \textit{(ii)} task tracking enables the teacher agent to generate high-dimensional tasks effectively, and \textit{(iii)} performance monitoring ensures the curriculum continually matches the student’s learning progress. Together, these elements result in more efficient and robust policy learning for real-world robot deployment.

While GACL demonstrates clear benefits, certain limitations warrant acknowledgment: the framework's dependence on pre-training the VAE with real-world task data may require substantial data collection in domains where examples are scarce, and our fixed grounding probability $\epsilon$ could benefit from adaptive scheduling based on learning progress.

In future work, GACL can be extended to a broader range of robotic tasks beyond navigation and locomotion, such as manipulation~\cite{haarnoja2018composable, kalashnikov2018scalable} or multi-agent coordination~\cite{limbu2023team, limbu2024scaling, zhouIROS2024, liu2021team}. Further research could focus on more sophisticated latent-space manipulation and curriculum design, enabling even richer task variations. Another promising direction involves transfer and lifelong learning~\cite{liu2021lifelong}, where pre-trained teacher agents in other domains accelerate adaptation to new ones. Finally, structured learning approaches---from cost-function shaping~\cite{xiao2022learning} and kinodynamic modeling~\cite{xiao2021learning, karnan2022vi, atreya2022high, pokhrel2024cahsor, datar2024learning, datar2024terrain} to planner parameter optimization~\cite{xiao2022appl, xiao2020appld, wang2021appli, wang2021apple, xu2021applr, das2024motion}---could further enhance GACL's efficiency and generalizability in real-world robotic systems.

\bibliographystyle{IEEEtran}
\bibliography{IEEEabrv,references}

\begin{thebibliography}{10}
\providecommand{\url}[1]{#1}
\csname url@samestyle\endcsname
\providecommand{\newblock}{\relax}
\providecommand{\bibinfo}[2]{#2}
\providecommand{\BIBentrySTDinterwordspacing}{\spaceskip=0pt\relax}
\providecommand{\BIBentryALTinterwordstretchfactor}{4}
\providecommand{\BIBentryALTinterwordspacing}{\spaceskip=\fontdimen2\font plus
\BIBentryALTinterwordstretchfactor\fontdimen3\font minus
  \fontdimen4\font\relax}
\providecommand{\BIBforeignlanguage}[2]{{%
\expandafter\ifx\csname l@#1\endcsname\relax
\typeout{** WARNING: IEEEtran.bst: No hyphenation pattern has been}%
\typeout{** loaded for the language `#1'. Using the pattern for}%
\typeout{** the default language instead.}%
\else
\language=\csname l@#1\endcsname
\fi
#2}}
\providecommand{\BIBdecl}{\relax}
\BIBdecl

\bibitem{xiao2024autonomous}
X.~Xiao, Z.~Xu, A.~Datar, G.~Warnell, P.~Stone, J.~J. Damanik, J.~Jung, C.~A.
  Deresa, T.~D. Huy, C.~Jinyu \emph{et~al.}, ``Autonomous ground navigation in
  highly constrained spaces: Lessons learned from the third barn challenge at
  icra 2024 [competitions],'' \emph{IEEE Robotics \& Automation Magazine},
  vol.~31, no.~3, pp. 197--204, 2024.

\bibitem{atanassov2024curriculum}
V.~Atanassov, J.~Ding, J.~Kober, I.~Havoutis, and C.~Della~Santina,
  ``Curriculum-based reinforcement learning for quadrupedal jumping: A
  reference-free design,'' \emph{arXiv preprint arXiv:2401.16337}, 2024.

\bibitem{portelas2020automatic}
R.~Portelas, C.~Colas, L.~Weng, K.~Hofmann, and P.-Y. Oudeyer, ``Automatic
  curriculum learning for deep rl: A short survey,'' \emph{arXiv preprint
  arXiv:2003.04664}, 2020.

\bibitem{dennis2020emergent}
\BIBentryALTinterwordspacing
M.~Dennis, N.~Jaques, R.~Turner, H.~Song, J.~Z. Leibo, E.~Hughes, and
  M.~Botvinick, ``Emergent complexity and zero-shot transfer via unsupervised
  environment design,'' \emph{arXiv preprint arXiv:2012.02096}, 2020. [Online].
  Available: \url{https://arxiv.org/abs/2012.02096}
\BIBentrySTDinterwordspacing

\bibitem{azad2023clutr}
A.~S. Azad, I.~Gur, J.~Emhoff, N.~Alexis, A.~Faust, P.~Abbeel, and I.~Stoica,
  ``Clutr: Curriculum learning via unsupervised task representation learning,''
  in \emph{International Conference on Machine Learning}.\hskip 1em plus 0.5em
  minus 0.4em\relax PMLR, 2023, pp. 1361--1395.

\bibitem{xu2023benchmarking}
Z.~Xu, B.~Liu, X.~Xiao, A.~Nair, and P.~Stone, ``Benchmarking reinforcement
  learning techniques for autonomous navigation,'' in \emph{2023 IEEE
  International Conference on Robotics and Automation (ICRA)}.\hskip 1em plus
  0.5em minus 0.4em\relax IEEE, 2023, pp. 9224--9230.

\bibitem{xu2024dexterous}
Z.~Xu, A.~H. Raj, X.~Xiao, and P.~Stone, ``Dexterous legged locomotion in
  confined 3d spaces with reinforcement learning,'' \emph{arXiv preprint
  arXiv:2403.03848}, 2024.

\bibitem{bengio2009curriculum}
Y.~Bengio, J.~Louradour, R.~Collobert, and J.~Weston, ``Curriculum learning,''
  in \emph{Proceedings of the 26th annual international conference on machine
  learning}, 2009, pp. 41--48.

\bibitem{perille2020benchmarking}
D.~Perille, A.~Truong, X.~Xiao, and P.~Stone, ``Benchmarking metric ground
  navigation,'' in \emph{2020 IEEE International Symposium on Safety, Security,
  and Rescue Robotics (SSRR)}.\hskip 1em plus 0.5em minus 0.4em\relax IEEE,
  2020, pp. 116--121.

\bibitem{xu2024reinforcement}
T.~Xu, C.~Pan, and X.~Xiao, ``Reinforcement learning for wheeled mobility on
  vertically challenging terrain,'' in \emph{2024 IEEE International Symposium
  on Safety Security Rescue Robotics (SSRR)}.\hskip 1em plus 0.5em minus
  0.4em\relax IEEE, 2024, pp. 125--130.

\bibitem{narvekar2020curriculum}
S.~Narvekar, B.~Peng, M.~Leonetti, J.~Sinapov, M.~E. Taylor, and P.~Stone,
  ``Curriculum learning for reinforcement learning domains: A framework and
  survey,'' \emph{Journal of Machine Learning Research}, vol.~21, no. 181, pp.
  1--50, 2020.

\bibitem{campbell2023automatic}
R.~Campbell and J.~Yoon, ``Automatic curriculum learning with gradient reward
  signals,'' \emph{arXiv preprint arXiv:2312.13565}, 2023.

\bibitem{kingma2013auto}
D.~P. Kingma, ``Auto-encoding variational bayes,'' \emph{arXiv preprint
  arXiv:1312.6114}, 2013.

\bibitem{schulman2017proximal}
J.~Schulman, F.~Wolski, P.~Dhariwal, A.~Radford, and O.~Klimov, ``Proximal
  policy optimization algorithms,'' \emph{arXiv preprint arXiv:1707.06347},
  2017.

\bibitem{makoviychuk2021isaac}
V.~Makoviychuk, L.~Wawrzyniak, Y.~Guo, M.~Lu, K.~Storey, M.~Macklin,
  D.~Hoeller, N.~Rudin, A.~Allshire, A.~Handa \emph{et~al.}, ``Isaac gym: High
  performance gpu-based physics simulation for robot learning,'' \emph{arXiv
  preprint arXiv:2108.10470}, 2021.

\bibitem{haarnoja2018composable}
T.~Haarnoja, V.~Pong, A.~Zhou, M.~Dalal, P.~Abbeel, and S.~Levine, ``Composable
  deep reinforcement learning for robotic manipulation,'' in \emph{2018 IEEE
  international conference on robotics and automation (ICRA)}.\hskip 1em plus
  0.5em minus 0.4em\relax IEEE, 2018, pp. 6244--6251.

\bibitem{kalashnikov2018scalable}
D.~Kalashnikov, A.~Irpan, P.~Pastor, J.~Ibarz, A.~Herzog, E.~Jang, D.~Quillen,
  E.~Holly, M.~Kalakrishnan, V.~Vanhoucke \emph{et~al.}, ``Scalable deep
  reinforcement learning for vision-based robotic manipulation,'' in
  \emph{Conference on robot learning}.\hskip 1em plus 0.5em minus 0.4em\relax
  PMLR, 2018, pp. 651--673.

\bibitem{limbu2023team}
M.~Limbu, Z.~Hu, S.~Oughourli, X.~Wang, X.~Xiao, and D.~Shishika, ``Team
  coordination on graphs with state-dependent edge costs,'' in \emph{2023
  IEEE/RSJ International Conference on Intelligent Robots and Systems
  (IROS)}.\hskip 1em plus 0.5em minus 0.4em\relax IEEE, 2023, pp. 679--684.

\bibitem{limbu2024scaling}
M.~Limbu, Z.~Hu, X.~Wang, D.~Shishika, and X.~Xiao, ``Scaling team coordination
  on graphs with reinforcement learning,'' in \emph{2024 IEEE International
  Conference on Robotics and Automation (ICRA)}, 2024, pp. 16\,538--16\,544.

\bibitem{zhouIROS2024}
Y.~Zhou, M.~Limbu, G.~J. Stein, X.~Wang, D.~Shishika, and X.~Xiao, ``Team
  coordination on graphs: Problem, analysis, and algorithms,'' in \emph{2024
  IEEE/RSJ International Conference on Intelligent Robots and Systems
  (IROS)}.\hskip 1em plus 0.5em minus 0.4em\relax IEEE, 2024.

\bibitem{liu2021team}
B.~Liu, X.~Xiao, and P.~Stone, ``Team orienteering coverage planning with
  uncertain reward,'' in \emph{2021 IEEE/RSJ International Conference on
  Intelligent Robots and Systems (IROS)}.\hskip 1em plus 0.5em minus
  0.4em\relax IEEE, 2021, pp. 9728--9733.

\bibitem{liu2021lifelong}
------, ``A lifelong learning approach to mobile robot navigation,'' \emph{IEEE
  Robotics and Automation Letters}, vol.~6, no.~2, pp. 1090--1096, 2021.

\bibitem{xiao2022learning}
X.~Xiao, T.~Zhang, K.~M. Choromanski, T.-W.~E. Lee, A.~Francis, J.~Varley,
  S.~Tu, S.~Singh, P.~Xu, F.~Xia, S.~M. Persson, L.~Takayama, R.~Frostig,
  J.~Tan, C.~Parada, and V.~Sindhwani, ``Learning model predictive controllers
  with real-time attention for real-world navigation,'' in \emph{Conference on
  robot learning}.\hskip 1em plus 0.5em minus 0.4em\relax PMLR, 2022.

\bibitem{xiao2021learning}
X.~Xiao, J.~Biswas, and P.~Stone, ``Learning inverse kinodynamics for accurate
  high-speed off-road navigation on unstructured terrain,'' \emph{IEEE Robotics
  and Automation Letters}, vol.~6, no.~3, pp. 6054--6060, 2021.

\bibitem{karnan2022vi}
H.~Karnan, K.~S. Sikand, P.~Atreya, S.~Rabiee, X.~Xiao, G.~Warnell, P.~Stone,
  and J.~Biswas, ``Vi-ikd: High-speed accurate off-road navigation using
  learned visual-inertial inverse kinodynamics,'' in \emph{2022 IEEE/RSJ
  International Conference on Intelligent Robots and Systems (IROS)}.\hskip 1em
  plus 0.5em minus 0.4em\relax IEEE, 2022, pp. 3294--3301.

\bibitem{atreya2022high}
P.~Atreya, H.~Karnan, K.~S. Sikand, X.~Xiao, S.~Rabiee, and J.~Biswas,
  ``High-speed accurate robot control using learned forward kinodynamics and
  non-linear least squares optimization,'' in \emph{2022 IEEE/RSJ International
  Conference on Intelligent Robots and Systems (IROS)}.\hskip 1em plus 0.5em
  minus 0.4em\relax IEEE, 2022, pp. 11\,789--11\,795.

\bibitem{pokhrel2024cahsor}
A.~Pokhrel, A.~Datar, M.~Nazeri, and X.~Xiao, ``{CAHSOR}: Competence-aware
  high-speed off-road ground navigation in {SE} (3),'' \emph{IEEE Robotics and
  Automation Letters}, 2024.

\bibitem{datar2024learning}
A.~Datar, C.~Pan, and X.~Xiao, ``Learning to model and plan for wheeled
  mobility on vertically challenging terrain,'' \emph{IEEE Robotics and
  Automation Letters}, 2024.

\bibitem{datar2024terrain}
A.~Datar, C.~Pan, M.~Nazeri, A.~Pokhrel, and X.~Xiao, ``Terrain-attentive
  learning for efficient 6-dof kinodynamic modeling on vertically challenging
  terrain,'' in \emph{2024 IEEE/RSJ International Conference on Intelligent
  Robots and Systems (IROS)}.\hskip 1em plus 0.5em minus 0.4em\relax IEEE,
  2024.

\bibitem{xiao2022appl}
X.~Xiao, Z.~Wang, Z.~Xu, B.~Liu, G.~Warnell, G.~Dhamankar, A.~Nair, and
  P.~Stone, ``Appl: Adaptive planner parameter learning,'' \emph{Robotics and
  Autonomous Systems}, vol. 154, p. 104132, 2022.

\bibitem{xiao2020appld}
X.~Xiao, B.~Liu, G.~Warnell, J.~Fink, and P.~Stone, ``Appld: Adaptive planner
  parameter learning from demonstration,'' \emph{IEEE Robotics and Automation
  Letters}, vol.~5, no.~3, pp. 4541--4547, 2020.

\bibitem{wang2021appli}
Z.~Wang, X.~Xiao, B.~Liu, G.~Warnell, and P.~Stone, ``Appli: Adaptive planner
  parameter learning from interventions,'' in \emph{2021 IEEE international
  conference on robotics and automation (ICRA)}.\hskip 1em plus 0.5em minus
  0.4em\relax IEEE, 2021, pp. 6079--6085.

\bibitem{wang2021apple}
Z.~Wang, X.~Xiao, G.~Warnell, and P.~Stone, ``Apple: Adaptive planner parameter
  learning from evaluative feedback,'' \emph{IEEE Robotics and Automation
  Letters}, vol.~6, no.~4, pp. 7744--7749, 2021.

\bibitem{xu2021applr}
Z.~Xu, G.~Dhamankar, A.~Nair, X.~Xiao, G.~Warnell, B.~Liu, Z.~Wang, and
  P.~Stone, ``Applr: Adaptive planner parameter learning from reinforcement,''
  in \emph{2021 IEEE international conference on robotics and automation
  (ICRA)}.\hskip 1em plus 0.5em minus 0.4em\relax IEEE, 2021, pp. 6086--6092.

\bibitem{das2024motion}
D.~Das, Y.~Lu, E.~Plaku, and X.~Xiao, ``Motion memory: Leveraging past
  experiences to accelerate future motion planning,'' in \emph{2024 IEEE
  International Conference on Robotics and Automation (ICRA)}.\hskip 1em plus
  0.5em minus 0.4em\relax IEEE, 2024, pp. 16\,467--16\,474.

\end{thebibliography}

\end{document}